\documentclass[10pt,twocolumn,letterpaper]{article}

\usepackage[pagenumbers]{cvpr} % To force page numbers, e.g. for an arXiv version

\usepackage{graphicx}
\usepackage{amsmath}
\usepackage{amssymb}
\usepackage{booktabs}
\usepackage{multicol}
\usepackage{multirow}
\usepackage[accsupp]{axessibility}

\usepackage[pagebackref,breaklinks,colorlinks]{hyperref}

\usepackage[capitalize]{cleveref}
\crefname{section}{Sec.}{Secs.}
\Crefname{section}{Section}{Sections}
\Crefname{table}{Table}{Tables}
\crefname{table}{Tab.}{Tabs.}

\def\cvprPaperID{155} % *** Enter the CVPR Paper ID here
\def\confName{CVPR}
\def\confYear{2023}

\begin{document}

%%%%%%%%% TITLE - PLEASE UPDATE
\title{MSINet: Twins Contrastive Search of Multi-Scale Interaction for Object ReID}

\author{Jianyang Gu$^{1}\quad$ Kai Wang$^{2}\quad$ Hao Luo$^{3}\quad$ Chen Chen$^{4}\quad$ Wei Jiang$^{1*}$\\ Yuqiang Fang$^{5}\quad$ Shanghang Zhang$^{6}\quad$ Yang You$^{2}\quad$ Jian Zhao$^{7}$\thanks{Co-corresponding authors} \\
$^{1}$Zhejiang University $\quad^{2}$National University of Singapore $\quad^{3}$Alibaba Group\\
$^{4}$OPPO Research Institute $\quad^{5}$Space Engineering University $\quad^{6}$Peking University\\
$^{7}$Institute of North Electronic Equipment \\
{\tt\small \{gu\_jianyang, jiangwei\_zju\}@zju.edu.cn zhaojian90@u.nus.edu}
% For a paper whose authors are all at the same institution,
% omit the following lines up until the closing ``}''.
% Additional authors and addresses can be added with ``\and'',
% just like the second author.
% To save space, use either the email address or home page, not both
% \and
% Second Author\\
% Institution2\\
% First line of institution2 address\\
% {\tt\small secondauthor@i2.org}
}
\maketitle

%%%%%%%%% ABSTRACT
\begin{abstract}
Neural Architecture Search (NAS) has been increasingly appealing to the society of object Re-Identification (ReID), for that task-specific architectures significantly improve the retrieval performance. Previous works explore new optimizing targets and search spaces for NAS ReID, yet they neglect the difference of training schemes between image classification and ReID. In this work, we propose a novel Twins Contrastive Mechanism (TCM) to provide more appropriate supervision for ReID architecture search. TCM reduces the category overlaps between the training and validation data, and assists NAS in simulating real-world ReID training schemes. We then design a Multi-Scale Interaction (MSI) search space to search for rational interaction operations between multi-scale features. In addition, we introduce a Spatial Alignment Module (SAM) to further enhance the attention consistency confronted with images from different sources. Under the proposed NAS scheme, a specific architecture is automatically searched, named as MSINet. Extensive experiments demonstrate that our method surpasses state-of-the-art ReID methods on both in-domain and cross-domain scenarios. Source code available in \href{https://github.com/vimar-gu/MSINet}{https://github.com/vimar-gu/MSINet}. 
\end{abstract}

%%%%%%%%% BODY TEXT
\section{Introduction}
\label{sec:intro}

Object re-identification (Re-ID) aims at retrieving specific object instances across different views~\cite{yi2014deep,zheng2016person,liu2016deep,liu2017provid,wang2021robust}, which attracts much attention in computer vision community due to its wide-range applications. Previous works have achieved great progresses on both supervised~\cite{luo2019strong,shen2017learning,wang2017orientation} and unsupervised ReID tasks~\cite{song2020unsupervised,ge2020self,zou2020joint}, most of which adopts backbone models originally designed for general image classification tasks~\cite{he2016deep,szegedy2016rethinking}. 

\begin{figure}
\centering
\includegraphics[width=0.48\textwidth]{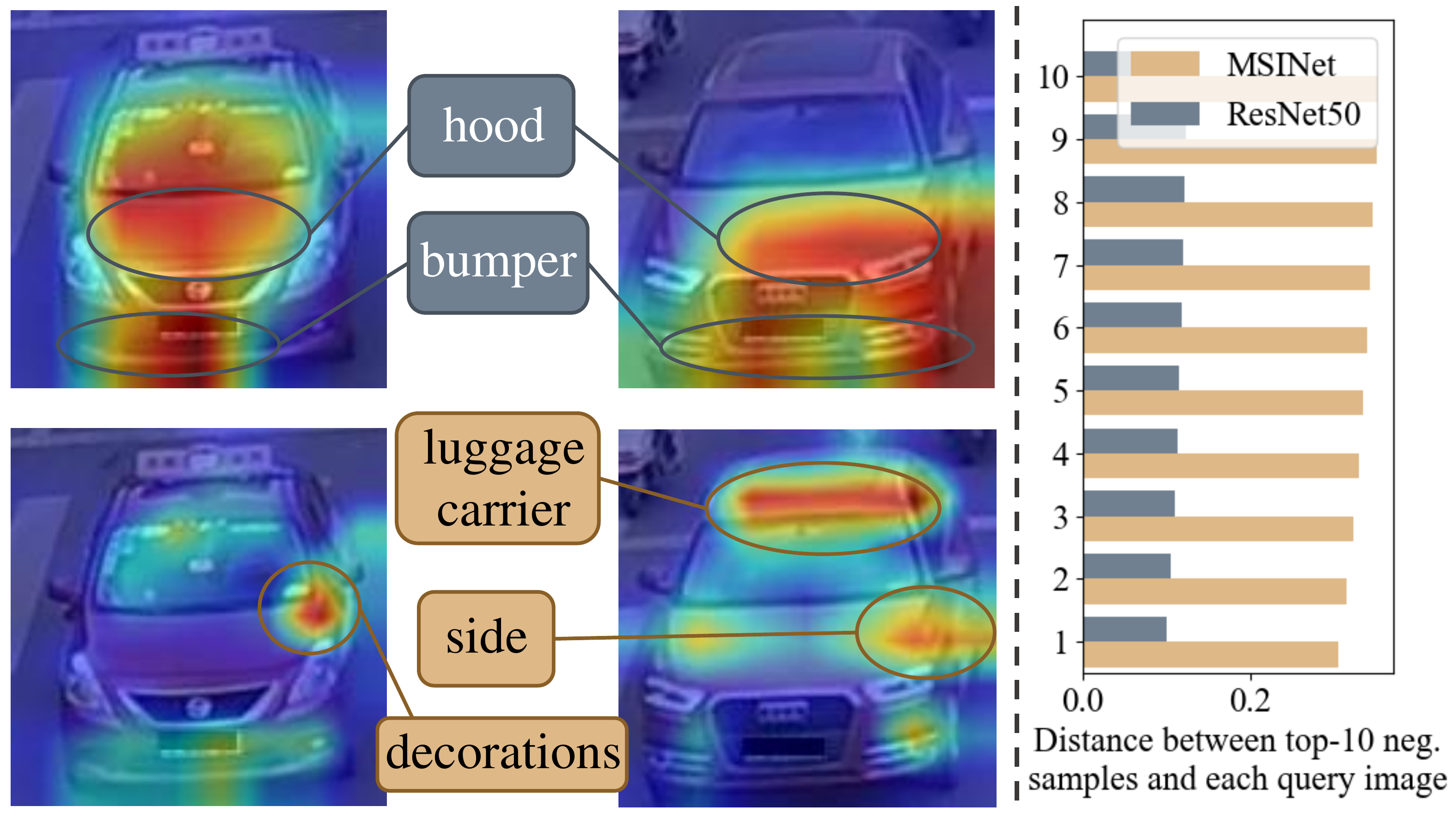}
\caption{The left panel shows the example activation maps of ResNet50 (1st row) and MSINet (2nd row). The right panel shows the average distances between the most similar 10 negative samples and each query image at the inference. Best viewed in color. }
\label{fig:intro}
\end{figure}

Recent literature~\cite{ye2021deep,zhou2019omni} has shown that applying different architectures on ReID leads to large performance variations. Some works employ Neural Architecture Search (NAS) for ReID~\cite{quan2019auto,li2021combined}. The proposed optimizing targets and search spaces stably improve the model performance, yet the main search scheme still follows traditional NAS methods designed for general classification tasks~\cite{liu2018darts,dong2019searching}. 
As an open-set task, ReID contains different categories in the training and validation sets~\cite{ye2021deep,zheng2017sift}, while the two sets share exactly the same categories in standard classification tasks~\cite{deng2009imagenet}, which is also followed by traditional NAS methods. 
The incompatibility between search schemes and real-world training schemes makes the searched architecture sub-optimal for ReID.
Moreover, ReID is required to distinguish more subtle distinctions among fine-grained instances compared with image-level classification~\cite{schroff2015facenet,yan2021beyond}. Some previous works~\cite{qian2017multi,zhou2019omni,zheng2020multi,chen2020orientation} have manifested that local perspectives and multi-scale features are discriminative for ReID. 
However, current utilizations of these features are mostly empirically designed, which can be more flexible according to the characteristics of different network layers. 

In this work, we propose a novel NAS scheme aiming at addressing the aforementioned challenges. 
In order to simulate the real-world ReID training schemes, a Twins Contrastive Mechanism (TCM) is proposed to unbind the categories of the training and validation sets. 
An adjustable overlap ratio of categories builds up the compatibility between NAS and ReID, which provides more appropriate supervision for ReID architecture search. 
Moreover, to search for more rational utilizations of multi-scale features, we design a Multi-Scale Interaction (MSI) search space. The MSI space focuses on interaction operations between multi-scale features along the shallow and deep layers of the network, which guides the features to promote each other.
Additionally, to further improve the generalization capability, we propose a Spatial Alignment Module (SAM) to enhance the attention consistency of the model confronted with images from different sources. 
With the above NAS scheme, we obtain a light-weight yet effective model architecture, denoted as Multi-Scale Interaction Net (MSINet). 

% We evaluate the proposed MSINet qualitatively and quantitatively on multiple ReID datasets. 
We visualize the example activation maps of our proposed MSINet and ResNet50~\cite{he2016deep} trained on VeRi-776~\cite{liu2016deep,liu2016large} in Fig.~\ref{fig:intro}. 
Compared to ResNet50, MSINet focuses on more unique distinctions with specific semantic information to recognize instances. 
Besides, MSINet largely increases the distance margin between query image and corresponding negative samples, reflecting extraordinary discriminative capability.
Extensive experiments demonstrate that MSINet surpasses state-of-the-art (SOTA) ReID methods on both in-domain and cross-domain scenarios. Our source codes are available in the supplementary material. 

Our contributions are summarized as follows:
\begin{itemize}
    \item To the best of our knowledge, we are the first to build the NAS search scheme according to the real-world ReID training schemes, which provides more appropriate supervision for the ReID architecture search. 
    \item We propose a novel search space based on the Multi-Scale Interaction (MSI) operations and a Spatial Alignment Module (SAM) to improve the model performance on in-domain and cross-domain scenarios. 
    \item We construct a light-weight yet effective architecture for ReID tasks, denoted as MSINet. With only 2.3M parameters, MSINet surpasses ResNet50~\cite{he2016deep} by 9\% mAP on MSMT17~\cite{wei2018person} and 16\% mAP on MSMT17$\to$Market-1501~\cite{zheng2015scalable}. 
\end{itemize}

\section{Related Works}

{\bf Neural Architecture Search.}
NAS has been increasingly appealing to the computer vision society, due to its automatic architecture designing characteristics. NAS methods can be roughly separated into four categories: reinforcement learning~\cite{zoph2018learning,bello2017neural}, evolutionary algorithms~\cite{real2019regularized,liu2018hierarchical}, gradient desent~\cite{liu2018darts,luo2018neural} and performance prediction~\cite{domhan2015speeding,liu2018progressive}. Liu \emph{et al.} establish a differentiable architecture search (DARTS) method~\cite{liu2018darts}, which improves the practicability of NAS by a large extent. Some later works further improve the structure through sampling strategy~\cite{xu2019pc}, network pruning~\cite{cai2018proxylessnas,dai2020data}, progressive learning~\cite{chen2019progressive}, collaborative competition~\cite{chu2020fair}, \emph{etc}. 
Most of NAS works focus on general image classification tasks, where the training and validation sets share the exact same categories. 
Following the setting, however, leads to incompatibility with the real-world training schemes of object ReID. 
In this work, we unbind the category bond between the two sets and propose a novel search scheme suitable for ReID. 

{\bf ReID Network Design.}
Current ReID works mostly adopt backbones designed for image classification~\cite{he2016deep,huang2017densely,szegedy2016rethinking,sandler2018mobilenetv2}. Some works~\cite{zhang2020relation,fang2019bilinear,li2018harmonious} design attention modules based on the common backbones to unearth their potential on distinguishing local distinctions. However, these methods usually lead to large calculation consumption. 

There are also several works focusing on designing ReID-specific architectures. 
Li \emph{et al.} present a Filter Pairing Neural Network to dynamically match patches in the feature maps~\cite{li2014deepreid}. Wang \emph{et al.} separate and regroup the features of two samples with a WConv layer~\cite{wang2018person}. Guo \emph{et al.} extract multi-scale features to directly evaluate the similarity between samples~\cite{guo2018efficient}. However, the siamese structure is inconvenient when conducting retrieval on large galleries. 
Zhou \emph{et al.} aggregate multi-scale information to achieve high accuracy with small computing consumption~\cite{zhou2019omni}. Quan \emph{et al.} introduce a part-aware module into the DARTS search space~\cite{liu2018darts,quan2019auto}. Li \emph{et al.} propose a new search space in regard to receptive field scales~\cite{li2021combined}. These methods have excellent performance on limited parameter scales, but fail to surpass those networks with complex structures. 
Different from previous works, we design a light-weight searching structure focusing on rational interaction operations between multi-scale features. 
The searched MSINet surpasses SOTA methods on both in-domain and cross-domain tasks. 
% Served as the backbone, it is also able to boost unsupervised ReID methods. 

\section{Methods}
Our goal is to construct an effective NAS scheme to search for a light-weight backbone architecture suitable for ReID tasks. Based on the training schemes of ReID, we propose a novel Twins Contrastive Mechanism to provide more appropriate supervision for the search process. Aiming at rational interaction between multi-scale features, we design a Multi-Scale Interaction search space. We further introduce a Spatial Alignment Module to improve the generalization capability with limited parameter growth. 

\begin{figure*}[t]
\centering
\includegraphics[width=0.95\textwidth]{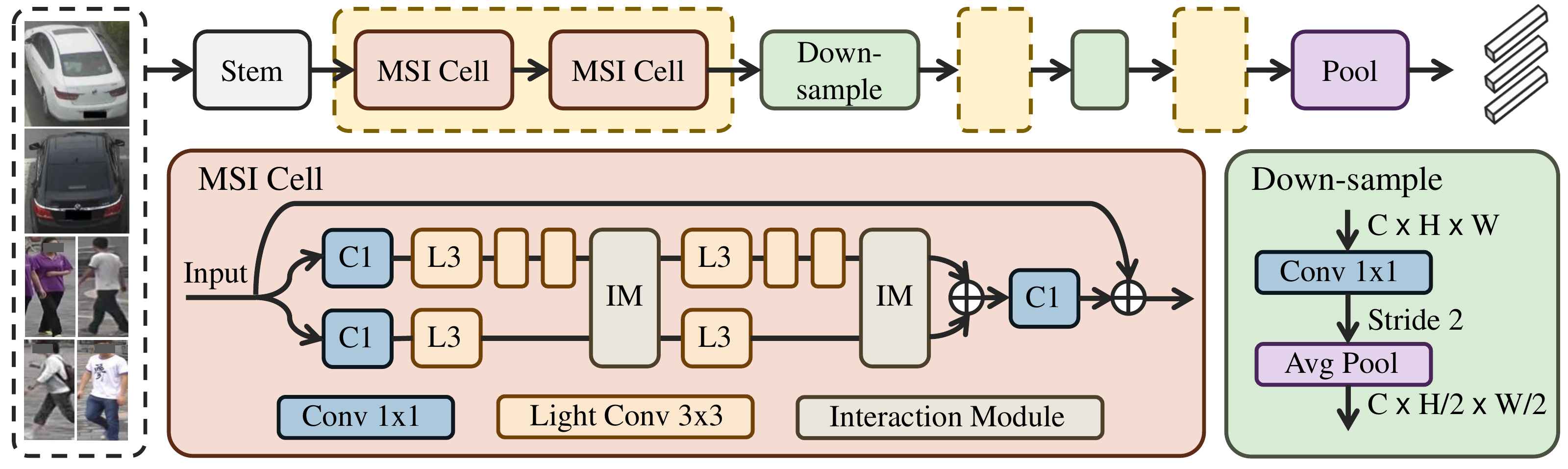}
\caption{The model structure of the proposed MSINet. The input can be either \emph{person} or \emph{vehicle}. Inside a cell, the input is separated to two branches, with different receptive field scales. The interaction module exchanges information between two branches. Architecture search automatically select the most appropriate interaction for each cell. }
\label{fig:structure}
\end{figure*}

\subsection{Twins Contrastive Mechanism}
\label{sec:memory}
NAS aims at automatically searching for the optimal network architecture for certain data.
Inspired by~\cite{liu2018darts}, a basic differentiable architecture search scheme is established. 
We define the ordinary model parameters as $\omega$, and architecture parameters as $\alpha$. For network layer $i$ with a search space of $\mathcal{O}$, $\alpha_{i}$ controls the weight of each operation $o$ in the space. The features are parallelly passed through all the operations, and the final output is formulated by the softmax-weighted sum of operation outputs:
\begin{equation}
\label{eq:darts}
    f(\mathbf{x}_i)=\sum_{o\in \mathcal{O}}{\frac{\exp{\left\{\alpha^o_i\right\}}}{\sum_{o^{'}\in \mathcal{O}}\exp{\left\{\alpha^{o^{'}}_i\right\}}} \cdot o(\mathbf{x}_i)}.
\end{equation}
The search process is conducted in an alternative manner. Training data is utilized to update the model parameters, and validation data is then employed to update the architecture parameters. For most NAS methods designed for image classification tasks, the training and validation data share exactly the same categories and a linear classification layer for loss calculation. 

Different from standard image classification, as an open-set retrieval task, ReID has different categories in the training and validation sets. 
The incompatibility between search schemes and real-world training schemes might lead to sub-optimal searching results. 
%Directly applying traditional NAS scheme might lead to sub-optimal searching results for real-world appliance. 
Accordingly, we propose a novel Twins Contrastive Mechanism (TCM) for NAS ReID training. 
Specifically, we employ two independent auxiliary memories $\mathcal{C}_{tr}$ and $\mathcal{C}_{val}$ to store the embedded features of the training and validation data, respectively. The memories are initialized with the centroid features, which are calculated by averaging the features of each category. 
At each iteration, the training loss is first calculated with $\mathcal{C}_{tr}$ for model parameter updating. Given an embedded feature $\mathbf{f}$ with category label $j$, the contrastive classification loss is calculated with:
\begin{equation}
\label{eq:clsloss}
    \mathcal{L}^{cls}_{tr}=-\log\frac{\exp(\mathbf{f}\cdot \mathbf{c}^j_{tr}/\tau)}{\sum^{N^c_{tr}}_{n=0}\exp(\mathbf{f}\cdot \mathbf{c}^n_{tr}/\tau)},
\end{equation}
where $\mathbf{c}^n_{tr}$ represents the memorized feature of category $n$, $N^c_{tr}$ stands for the total number of categories in the training set, and $\tau$ is the temperature parameter, which is set as 0.05 empirically~\cite{ge2020self}.
After updating the model parameters, the embedded feature $\mathbf{f}$ with category label $j$ is integrated into the corresponding memorized feature $\mathbf{c}^j_{tr}$ by:
\begin{equation}
    \mathbf{c}^j_{tr} \leftarrow \beta \mathbf{c}^j_{tr} + (1-\beta) \mathbf{f},
\end{equation}
where $\beta$ is set as 0.2 empirically~\cite{ge2020self}.
Then the updated model is evaluated on the validation data to generate to validation loss with $\mathcal{C}_{val}$ replacing $\mathcal{C}_{tr}$ in Eq.~\ref{eq:clsloss}. 
The architecture parameter is then updated with the validation loss to finish an iteration. 

As the loss calculation does not rely on the linear classification layer, the categories of the training and validation sets are unbound. We are able to dynamically adjust the category overlap ratio in these two sets. 
The advantages of a proper overlap ratio are summarized as two folds. 
Firstly, TCM better simulates the real-world training of ReID and helps the model focus on truly discriminative distinctions. The differences between the training and validation data improves the generalization capability of the model. 
Secondly, a relatively small proportion of overlapped categories stabilizes the architecture parameter update through a consistent optimizing target with the model parameter update. 

\begin{figure*}[t]
\centering
\includegraphics[width=0.95\textwidth]{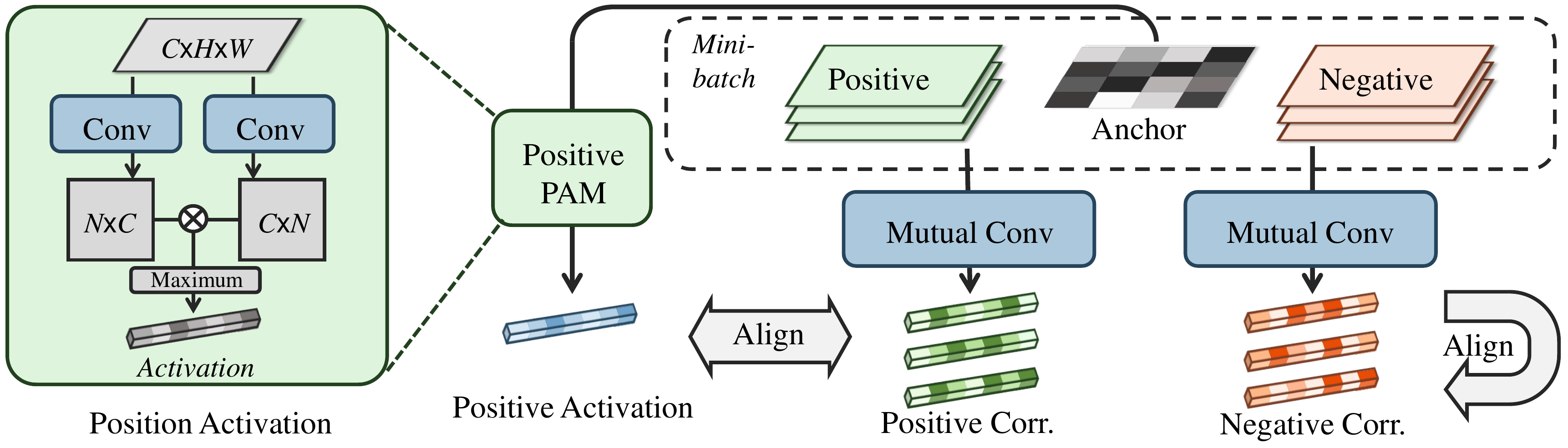}
\caption{The aligning pipeline of the proposed Spatial Alignment Module. The correlation activation vectors are calculated between the anchor feature and all features in a mini-batch. The positive vectors are aligned with the learnable self-activation, and the negative vectors are aligned with each other. The structure of PAM is shown on the left. }
\label{fig:sam}
\end{figure*}

\subsection{Multi-Scale Interaction Space}
\label{sec:space}
Although the local perspective and multi-scale features have already been investigated in previous ReID works~\cite{qian2017multi,zhou2019omni,li2021combined,sun2018beyond,zheng2020multi,chen2020orientation,zhou2019discriminative,zheng2019re}, the utilization of these information is mainly empirically designed aggregation, which is monotonous and restrained. 
We argue that on the one hand, the rational utilization of multi-scale features should be dynamically adjusted along the shallow and deep layers of the network. 
On the other hand, introducing interaction other than aggregation creates direct information exchange, and makes fuller use of multi-scale features. 
Therefore, we propose a novel Multi-Scale Interaction (MSI) search space to establish a light-weight architecture suitable for ReID. 

As shown in Fig.~\ref{fig:structure}, the network is mainly grouped with MSI cells and down-sample blocks, which is generally consistent with OSNet~\cite{zhou2019omni}. In each cell, the input features are passed through two branches with different receptive field scales. 
To reduce the calculation burden of the network, for the layers inside each branch, we adopt the stack of 1$\times$1 convolution and multiple depth-wise 3$\times$3 convolution to implement specific scales. A scale ratio $\rho$ of 3:1 is selected for the two branches. 
These two branches do not share model parameters, except for the Interaction Modules (IM). IM introduces information exchange for the two branches. There are 4 operation options for the IM. 
With the two-branch input features defined as $(\mathbf{x}_1, \mathbf{x}_2)$, the operations can be formulated as:

{\bf None.} None operation involves no parameters, and outputs exactly the input features $(\mathbf{x}_1, \mathbf{x}_2)$. 

{\bf Exchange.} Exchange acts as the strongest interaction among all options. It directly exchanges the features for the two branches and outputs $(\mathbf{x}_2, \mathbf{x}_1)$. 
Exchange contains no extra parameters, as well. 

\begin{table}[t]
\centering
\small
\caption{The detailed interaction operation in the proposed MSINet architecture. N: None; E: Exchange; G: Channel Gate; A: Cross Attention. }
\label{table:structure}
\begin{tabular}{c|c|c|c|c|c|c|c|c|c|c|c}
\toprule
    \multicolumn{2}{c|}{Cell \#1} & \multicolumn{2}{c|}{Cell \#2} & \multicolumn{2}{c|}{Cell \#3} & \multicolumn{2}{c|}{Cell \#4} & \multicolumn{2}{c|}{Cell \#5} & \multicolumn{2}{c}{Cell \#6} \\
    1 & 2 & 3 & 4 & 5 & 6 & 7 & 8 & 9 & 10 & 11 & 12 \\
    G & G & E & G & A & G & G & N & G & A & E & A \\
    \bottomrule
\end{tabular}
\end{table}

\begin{figure}[t]
    \centering
    \includegraphics[width=0.5\textwidth]{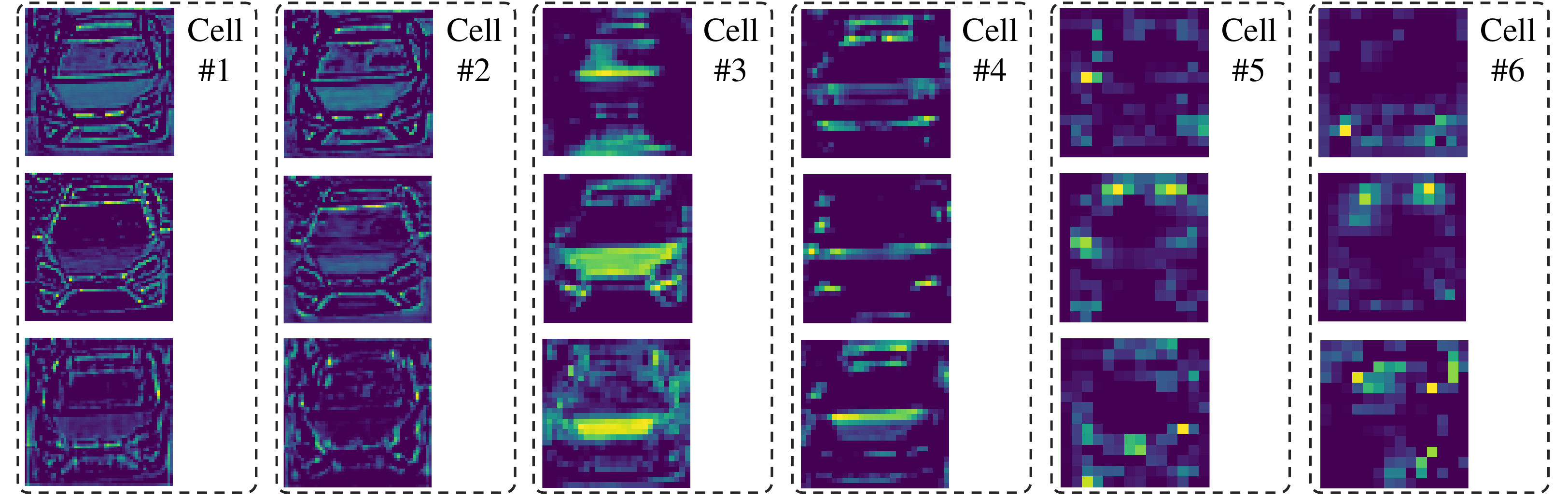}
    \caption{Output feature maps of MSI cells at different layers. }
    \label{fig:filter}
\end{figure}

{\bf Channel Gate.} Channel gate introduces a Multi-Layer Perceptron (MLP) to generate a channel-wise attention gate~\cite{zhou2019omni,woo2018cbam} as:
\begin{equation}
    G(\mathbf{x})=\sigma(MLP(\mathbf{x}))),
\end{equation}
and returns $(G(\mathbf{x}_1)\cdot \mathbf{x}_1, G(\mathbf{x}_2)\cdot \mathbf{x}_2)$. 
The MLP is composed of 2 fully connected layers and its parameters are shared for both branches. Thereby it achieves interaction by jointly screening discriminative feature channels. 

{\bf Cross Attention.} 
Traditional channel attention module calculates the channel correlation inside a single feature map\cite{fu2019dual}. The original feature map $\mathbf{x}\in \mathcal{R}^{C\times H\times W}$ is firstly reshaped into the \textit{query} feature $\tilde{\mathbf{x}}\in \mathcal{R}^{C\times N}$, where $N=H\times W$. Then the correlation activation is calculated by performing a matrix multiplication between the \textit{query} feature $\tilde{\mathbf{x}}$ and the \textit{key} feature $\tilde{\mathbf{x}}^\top$. We propose to exchange the \textit{keys} of the two branches to explicitly calculate the correlation between each other. The correlation activation is then transformed to a mask, and is added up to the original features with a learnable proportion. 

After interaction, the multi-scale branches are fused through a sum operation. 
It is worth noting that the extra parameters brought by multiple interaction options are limited, which enables searching for each cell along the whole network independently. 
At the beginning of the network, we employ the same stem module as that in OSNet~\cite{zhou2019omni}, containing a $7\times 7$ convolutional layer and a $3\times 3$ max pooling with a stride of 2. 
After the searching process, the interaction operation $o$ with the largest weight $\alpha_i^o$ at each layer is reserved to form the searched architecture. 

\begin{table*}[t]
\small
\begin{center}
\caption{Supervised performance on object ReID datsets. The results in the top part are trained from scratch, and those in the bottom part are pre-trained on ImageNet in advance. As the compared methods are originally proposed for person ReID, we reproduce the results in vehicle datasets. $^*$ indicates that the results of person ReID are reproduced by us. The evaluation results of architecture searched on VR can be found in the supplementary material. }
\label{table:supervised}
\setlength{\tabcolsep}{4pt}
\begin{tabular}{lcccccccccccccc}
    \toprule
    \multirow{2}{*}{Method} & \multirow{2}{*}{Params} & Inference & \multicolumn{2}{c}{M} & \multicolumn{2}{c}{MS} & \multicolumn{2}{c}{VR} & \multicolumn{2}{c}{VID} & \multicolumn{2}{c}{MS$\to$M} & \multicolumn{2}{c}{VR$\to$VID} \\
     & & Time & R-1$\uparrow$ & mAP$\uparrow$ & R-1$\uparrow$ & mAP$\uparrow$ & R-1$\uparrow$ & mAP$\uparrow$ & R-1$\uparrow$ & R-5$\uparrow$ & R-1$\uparrow$ & mAP$\uparrow$ & R-1$\uparrow$ & R-5$\uparrow$ \\
     \midrule
    ResNet50$^*$~\cite{luo2019strong} & $\sim$ 24M & 1$\times$ & 85.7 & 68.3 & 48.0 & 25.7 & 92.8 & 69.9 & 70.6 & 76.6 & - & - \\
    OSNet~\cite{zhou2019omni} & 2.2M & 0.79$\times$ & 93.6 & 81.0 & 71.0 & 43.3 & 95.4 & 72.8 & 76.0 & 88.7 & - & - & - & - \\
    CDNet~\cite{li2021combined} & 1.8M & 0.67$\times$ & 93.7 & 83.7 & 73.7 & 48.5 & 94.3 & 73.0 & 74.5 & 88.8 & - & - & - & - \\
    MSINet & 2.3M & 0.71$\times$ & \textbf{94.6} & \textbf{87.0} & \textbf{76.0} & \textbf{52.5} & \textbf{95.9} & \textbf{75.0} & \textbf{76.5} & \textbf{89.8} & - & - & - & - \\
    \midrule
    ResNet50$^*$~\cite{luo2019strong} & $\sim$24M & 1$\times$ & 94.5 & 85.9 & 75.5 & 50.4 & 94.5 & 73.6 & 76.5 & 89.9 & 58.8 & 31.8 & 42.8 & 61.9 \\
    OSNet~\cite{zhou2019omni} & 2.2M & 0.79$\times$ & 94.8 & 84.9 & 78.7 & 52.9 & 95.5 & 76.4 & 76.0 & 88.6 & 66.6 & 37.5 & 46.5 & 63.1 \\
    CDNet~\cite{li2021combined} & 1.8M & 0.67$\times$ & 95.1 & 86.0 & 78.9 & 54.7 & - & - & - & - & - & - & - & - \\
    MSINet & 2.3M & 0.71$\times$ & 95.3 & 89.6 & \textbf{81.0} & \textbf{59.6} & \textbf{96.8} & 78.8 & 77.9 & 91.7 & 74.9 & 46.2 & 48.0 & 65.6 \\
    MSINet-SAM & 2.4M & 0.71$\times$ & \textbf{95.5} & \textbf{89.9} & 80.7 & 59.5 & 96.7 & \textbf{79.0} & \textbf{78.0} & \textbf{91.9} & \textbf{76.3} & \textbf{48.4} & \textbf{49.0} & \textbf{66.8} \\
    \bottomrule
\end{tabular}
\end{center}
\end{table*}

After searching the architecture, the model is validated on various Re-ID tasks. 
The training is constrained by the classification id loss and the triplet loss, formulated by:
\begin{equation}
    \mathcal{L}_{id}=\frac{1}{N}\sum_{i=1}^N-\log\left(\frac{\exp\mathbf{W}^\top_i\mathbf{f}_i}{\sum_j\exp\mathbf{W}^\top_j\mathbf{f}_i}\right),
\end{equation}
where $\mathbf{f}_i$ is a feature vector, the corresponding classifier weight of which is $\mathbf{W}_i$, and
\begin{equation}
\label{eq:triplet}
    \mathcal{L}_{tri}=\left[\mathcal{D}(\mathbf{f}_a,\mathbf{f}_p)-\mathcal{D}(\mathbf{f}_a,\mathbf{f}_n)+m\right]_+,
\end{equation}
where $\mathbf{f}_a$, $\mathbf{f}_p$, $\mathbf{f}_n$ are the embedded features for the anchor, the hardest positive and negative samples in a mini-batch, $\mathcal{D}(\cdot,\cdot)$ is the Euclidean distance, $m$ is the margin parameter, and $[\cdot]_+$ is the $\max(\cdot,0)$ function. 

\subsection{Spatial Alignment Module}
\label{sec:sam}
The retrieval precision of object ReID tasks are largely affected by the variation of appearances such as poses, illumination and occlusion when the camera conditions change. In order that the model correctly and consistently focuses on the discriminative spatial positions, we design a Spatial Alignment Module (SAM) to explicitly align the spatial attention between images, as shown in Fig.~\ref{fig:sam}.  

Specifically, we first calculate the position-wise correlation activation map $\mathbf{A}$ between the feature maps in a mini-batch. The activation between sample $i$ and $j$ can be formulated as: $\mathbf{A}(i, j)=\tilde{\mathbf{x}}_j^\top\times \tilde{\mathbf{x}}_i$, where $\tilde{\mathbf{x}}\in \mathcal{R}^{C\times N}$ is reshaped from the original feature $\mathbf{x}\in \mathcal{R}^{C\times H\times W}$. Then we take the maximum activation for each position of sample $i$ as:
\begin{equation}
    \mathbf{a}(i,j)=\max_{dim=1}\mathbf{A}(i,j).
\end{equation}
The above process is denoted as ``Mutual Conv'' in Fig.~\ref{fig:sam}. 
We evaluate the consistency between activation vectors with cosine similarity. 
For negative samples specifically, there can be many different hints for recognition, some of which might be inappropriate, such as the backgrounds. 
By aligning all the correlations for sample $i$, we hope that the network can correct some attention bias and consistently focus on discriminative positions. 

However, through aligning positive sample pairs, the ID-related features are expected to be emphasized, which cannot be achieved by aligning negative pairs. 
Therefore, we introduce an extra position activation module (PAM) to generate supervision for the alignment between positive pairs. 
The spatial alignment loss is formulated as:
\begin{equation}
\begin{split}
    \mathcal{L}_{sa}(i)=\frac{1}{N_+}\sum_{p\in \mathcal{I}_+} \left(1-S(\hat{\mathbf{a}}(i), \mathbf{a}(i,p))\right)+\\
    \frac{1}{N_-}\sum_{n_1,n_2\in \mathcal{I}_-} \left(1-S(\mathbf{a}(i,n_1), \mathbf{a}(i,n_2))\right),
\end{split}
\end{equation}
where $\mathcal{I}_+$ contains positive indices for sample $i$, the total number of which is $N_+$, and vice versa. $\hat{\mathbf{a}}(i)$ stands for the generated activation vector for positive sample alignment, and $S(\cdot,\cdot)$ is the cosine similarity. 

% Traditional position attention module involves a softmax operation on the activation to generate the final attention. We discard the softmax operation as it suppresses activation with low values, leading to loss of information. 
% The alignment process is only conducted during the training phase to integrate attention consistency into the backbone network, and the module is discarded during testing. 

\section{Experiments}

\subsection{Datasets and Evaluation Metrics}
Our proposed method is evaluated on two person ReID datasets Market-1501~\cite{zheng2015scalable}, MSMT17~\cite{wei2018person}, and two vehicle ReID datasets VeRi-776~\cite{liu2016large,liu2016deep} and VehicleID~\cite{liu2016vid}. 
For simplicity, the four datasets are denoted as M, MS, VR and VID in the following sections, respectively. 
Evaluation metrics include Cumulative Matching Characteristic (CMC) and mean average precision (mAP), which are commonly utilized on ReID tasks. 

\subsection{Architecture Search}

We conduct the searching process on MSMT17. SGD is adopted for model parameter update with an initial learning rate of 0.025. The model is trained for 350 epochs in total. We adopt a warm-up strategy for the first 10 epochs. Then the learning rate is decayed by 0.1 at 150, 225 and 300 epochs, respectively. Adam~\cite{kingma2014adam} is adopted for the architecture parameter update with an initial learning rate of 0.002. The learning rate is decayed at the same pace. The images are reshaped to 256$\times$128 for person and 256$\times$256 for vehicles. Data augmentation includes random flip, random crop and random erasing~\cite{zhong2020random}. The searched architecture is presented in Tab.~\ref{table:structure}. The ``MSINet'' in the following experiment sections refers to this architecture. 

We visualize the feature maps extracted by each MSI cell in Fig.~\ref{fig:filter}. At the shallow layers of the network, the kernels mainly focus on overall contour information. Channel gate helps to filter out inferior information, such as the background. 
As we approach deeper layers, the extracted features each have specific semantic information, where cross attention is more likely to be selected for the interaction. It indicates that cross attention is more rational for exchanging high-level semantic information. 

\subsection{Comparison with Other Backbones}

\setlength{\tabcolsep}{3.6pt}
\begin{table}[t]
\centering
\small
\caption{Supervised performance comparison between MSINet and SOTA methods on M and MS datasets. }
\label{table:sota-person}
\centering
\begin{tabular}{lccccc}
\toprule
    \multirow{2}{*}{Method} & \multirow{2}{*}{} & \multicolumn{2}{c}{M} & \multicolumn{2}{c}{MS} \\
    && R-1$\uparrow$ & mAP$\uparrow$ & R-1$\uparrow$ & mAP$\uparrow$ \\
    \midrule
    PCB\cite{sun2018beyond} && 93.8 & 81.6 & 68.2 & 40.4 \\
    MGN\cite{wang2018learning} && 95.7 & 86.9 & 76.9 & 52.1 \\
    OSNet\cite{zhou2019omni} && 93.6 & 81.0 & 71.0 & 43.3 \\
    IANet\cite{hou2019interaction} && 94.4 & 83.1 & 75.5 & 46.8 \\
    DGNet\cite{zheng2019joint} && 94.8 & 86.0 & 77.2 & 52.3 \\
    Auto-ReID\cite{quan2019auto} && 94.5 & 85.1 & - & - \\
    SAN\cite{jin2020semantics} && \textbf{96.1} & 88.0 & 79.2 & 55.7 \\
    CDNet\cite{li2021combined} && 95.1 & 86.0 & 78.9 & 54.7 \\
    BAT-Net\cite{fang2019bilinear} && 95.1 & 87.4 & 79.5 & 56.8 \\
    SFT\cite{luo2019spectral} && 94.1 & 87.5 & 79.0 & 58.3 \\
    CTF\cite{zhang2021coarse} && 94.8 & 87.7 & - & - \\
    RGA-SC\cite{zhang2020relation} && \textbf{96.1} & 88.4 & 80.3 & 57.5 \\
    \textbf{MSINet} && 95.3 & \textbf{89.6} & \textbf{81.0} & \textbf{59.6} \\
    \bottomrule
\end{tabular}
\end{table}

\begin{table}[t]
\caption{Unsupervised performance applying MSINet to SOTA methods for USL on M and UDA on M$\to$MS. }
\label{table:sota-unsup}
\centering
\small
\begin{tabular}{lcccc}
\toprule
    \multirow{2}{*}{Method} & \multicolumn{2}{c}{M} & \multicolumn{2}{c}{M$\to$MS} \\
    & R-1$\uparrow$ & mAP$\uparrow$ & R-1$\uparrow$ & mAP$\uparrow$ \\
    \midrule
    MMCL\cite{wang2020unsupervised} & 80.3 & 45.5 & 40.8 & 15.1 \\
    % NMRT\cite{zhao2020unsupervised} & - & - & 43.7 & 19.8 \\
    MMT\cite{ge2019mutual} & - & - & 50.1 & 24.0 \\
    JVTC+\cite{li2020joint} & 79.5 & 47.5 & 48.6 & 25.1 \\
    CycAs\cite{wang2020cycas} & 84.8 & 64.8 & - & - \\
    GCL\cite{chen2021joint} & 87.3 & 66.8 & 51.1 & 27.0 \\
    MPRD\cite{ji2021meta} & 83.0 & 51.0 & - & - \\
    SpCL\cite{ge2020self} & 88.1 & 73.1 & 53.7 & 26.8 \\
    GS\cite{han2021group} & 92.3 & 79.2 & - & - \\
    \textbf{GS+MSINet} & 91.7 & 81.5 & - & - \\
    HDCRL\cite{cheng2022hybrid} & 92.4 & 81.7 & - & - \\
    \textbf{HDCRL+MSINet} & \textbf{92.9} & \textbf{82.7} & - & - \\
    IDM\cite{dai2021idm} & - & - & 61.3 & 33.5 \\
    \textbf{IDM+MSINet} & - & - & \textbf{66.0} & \textbf{37.8} \\
    \bottomrule
\end{tabular}
\end{table}

We first compare our proposed MSINet with ResNet50 and recent proposed light-weight backbones in both in-domain and cross-domain ReID scenarios. 

{\bf In-Domain Tasks.} 
We adopt a two-group supervised evaluation scheme similar to that in~\cite{zhou2019omni,li2021combined}: training from scratch and fine-tuning ImageNet\cite{deng2009imagenet} pre-trained models. The training parameters for both schemes are kept the same as that in architecture search, except for an initial learning rate of 0.065. Triplet loss and cross entropy loss are adopted for the parameter update. 
The margin $m$ in Eq.~\ref{eq:triplet} is set as 0.3. 
\cite{li2021combined} adopts an FBLNeck. We also employ the same structure. The results are shown in Tab.~\ref{table:supervised}. 

ResNet50 is the most commonly utilized backbone network in ReID tasks, yet holds the worst performance. 
Moreover, ResNet50 largely depends on ImageNet pre-training, while MSINet without pre-training has already surpassed pre-trained ResNet50 on all metrics. 
Compared with the other datasets, MS contains more variations on illumination, background and camera pose, and brings a large performance gap between ResNet50 and other methods. It also validates the inadequacy of image classification networks on ReID tasks. 
OSNet~\cite{zhou2019omni} and CDNet~\cite{li2021combined} are recently proposed architectures designed specifically for ReID tasks. Both architectures focus on fusing multi-scale features to better suit ReID. CDNet employs a traditional NAS scheme to search for the proper receptive field scales for each cell. MSINet fixes the receptive field scale and instead selects optimal interaction operations inside each cell. With only a bit more parameters, MSINet surpasses all the other backbones by a large margin. 
% , which verifies the effectiveness of interaction operations. 

\begin{figure*}[t]
    \centering
    \includegraphics[width=0.98\textwidth]{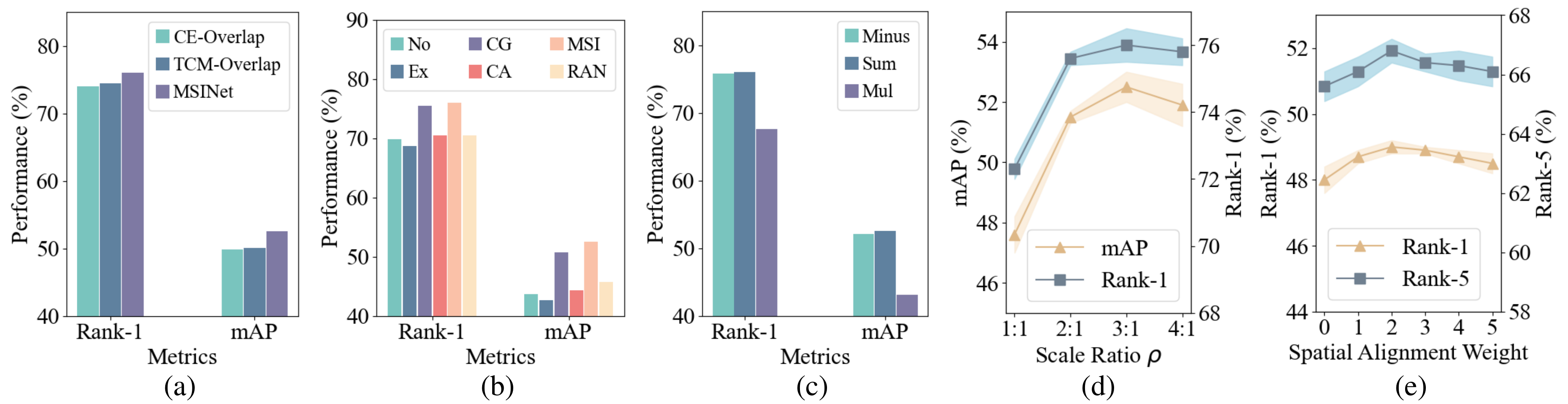}
    \caption{Ablation studies on (a) search scheme; (b) architecture; (c) multi-scale aggregation; (d) scale ratio $\rho$; (e) spatial alignment weight. }
    \label{fig:ablation}
\end{figure*}

\begin{table}[t]
\centering
\small
\caption{The effectiveness of each components in SAM. Both in-domain performance on VR and cross-domain performance on VR$\to$VID is evaluated.  }
\label{table:sam}
% \rowcolors{3}{lightgray!50}{white}
\begin{tabular}{lccccccc}
\toprule
    \multirow{2}{*}{Model} & \multirow{2}{*}{Pos.} & \multirow{2}{*}{Neg.} & \multirow{2}{*}{Align} & \multicolumn{2}{c}{VR} & \multicolumn{2}{c}{VR$\to$VID} \\
    & & & & R-1$\uparrow$ & mAP$\uparrow$ & R-1$\uparrow$ & R-5$\uparrow$ \\
    \midrule
    \multirow{6}{*}{MSINet} &  &  & - & 96.7 & 78.5 & 48.0 & 65.6 \\
    &\checkmark &  & Self & 96.7 & 78.1 & 48.4 & 66.0 \\
    & & \checkmark & Self & 96.7 & 78.4 & 48.5 & 66.2 \\
    &\checkmark & \checkmark & Unified & 96.6 & 78.3 & 48.3 & 65.8 \\
    &\checkmark & \checkmark & Separated & \textbf{96.8} & 78.6 & 48.7 & 66.4 \\
    &\checkmark & \checkmark & PAM-Self & 96.7 & \textbf{79.0} & \textbf{49.0} & \textbf{66.8} \\
    \midrule
    \multirow{2}{*}{OSNet}& & & None & 95.5 & 76.4 & 46.5 & 63.1 \\
    &\checkmark & \checkmark & PAM-Self & 95.9 & 76.3 & 47.5 & 63.3 \\
    \bottomrule
\end{tabular}
\end{table}

{\bf Cross-Domain Tasks.}
Cross-domain experiments verify the generalization capability of the model. Following previous domain generalizable ReID works~\cite{jin2020style,liao2020interpretable}, data augmentation is adjusted to random flip, random crop and color jittering. The model is pre-trained and fine-tuned for 250 epochs to avoid over-fitting. The other settings are kept the same as supervision scenes. With no present pre-trained models for CDNet~\cite{li2021combined}, it is excluded from this section. 

Tab.~\ref{table:supervised} shows that ResNet50 can be easily interfered by different image styles confronted with new image domains. 
OSNet learns multi-scale features with specific semantic information for ReID, which is domain invariant to some extent. 
Our proposed search scheme also takes into account the generalization capability of the model. By partly separating the categories for training and validation sets, the searched interaction operations generalize well confronted with new image domains. Except for discrimination, MSINet also surpasses the other backbones on cross-domain tasks by a large margin with faster inference speed. 

Additionally, we introduce SAM into the model, which aligns the spatial correlations between images. A weighted sum of ReID loss and spatial alignment loss is utilized when training the network with SAM. The weight of spatial alignment loss is set as $\lambda_{sa}=2.0$. Without extra inference consumption or damages on the supervised performance, SAM further boosts the generalization capability of MSINet. 

% \begin{figure*}[t]
%     \centering
%     \hskip -8pt
%     \subfigure[]{
%         \includegraphics[width=0.24\textwidth]{model_design.png}
%     }
%     \hskip -4pt
%     \subfigure[]{
%         \includegraphics[width=0.24\textwidth]{agg.png}
%     }
%     \hskip -4pt
%     \subfigure[]{
%         \includegraphics[width=0.24\textwidth]{branch_ratio.png}
%     }
%     \hskip -4pt
%     \subfigure[]{
%         \includegraphics[width=0.24\textwidth]{sa_weight.png}
%     }
%     \hskip -8pt
%     \caption{Ablation studies on (a) architecture search; (b) multi-scale aggregation; (c) scale ratio $\rho$; (d) spatial alignment weight. 
%     \label{fig:ablation}}
% \end{figure*}

\subsection{Comparison with State-of-the-art Methods}

Tab.~\ref{table:sota-person} further illustrates the supervision performance comparison of our proposed MSINet with the SOTA methods on M and MS datasets. 
With much less parameters than most of the compared methods, MSINet achieves a retrieval accuracy comparable to that of more complicated ones. 
Auto-ReID~\cite{quan2019auto} first designs a NAS scheme for ReID, yet the DARTS-style architecture contains 13M parameters. 
RGA-SC~\cite{zhang2020relation} carefully designs a relation-aware global attention module. 
MSINet achieves even higher performance with less training consumption, which validates the superiority of selecting rational interaction. 

We also evaluate the model performance replacing the backbone network from ResNet50 to MSINet for SOTA unsupervised ReID methods in Tab.~\ref{table:sota-unsup}. 
For purely unsupervised learning (USL) method GS~\cite{han2021group} on M dataset, MSINet performs slightly lower on rank-1, yet has a large superiority on mAP. 
For HDCRL~\cite{cheng2022hybrid}, MSINet shows obvious superiority over ResNet50. 
For unsupervised domain adaptation (UDA) method IDM~\cite{dai2021idm} on M$\to$MS task, MSINet surpasses ResNet50 by a large margin, which further proves that the TCM brings outstanding generalization capability to the searched architecture. 

\subsection{Ablation Studies}

{\bf Effectiveness of Architecture Search.}
To verify the effectiveness, we conduct supervised training on MS with different search schemes in Fig.~\ref{fig:ablation} (a). 
Under the standard classification scheme (``CE Overlap''), the searched model performs poor. Replacing the cross entropy loss to contrastive loss (``TCM Overlap'') only brings a slight improvement. The complete TCM framework unbounds the categories between the training and validation set, and thereby improves the performance by a large margin. 
% ``Overlap'' stands for sharing identical categories between the training and validation sets. TCM brings a slight improvement, while unbinding the categories better simulates the real-world training of ReID, and shows distinct advantages over the traditional cross entropy (CE) scheme. 

In Fig.~\ref{fig:ablation} (b) we compare the performance of different architectures. 
Firstly, we validate 4 models each with a unique interaction operation from the 4 options in the search space. 
None and Exchange, with no trainable parameters, achieve poor performance. Channel gate introduces channel-wise attention, whose model performs the best among 4 options. Cross attention exchanges the key features for the two branches. Over-frequent exchange interferes the ordinary feature extraction and degrades the network performance. 
Through appropriately arranging interaction operations along the architecture, MSINet surpasses all the above 4 models. Random architecture, on the other hand, shows no rational appliance of interaction operations, which validates that the proposed search scheme helps find suitable architectures for ReID. 

{\bf Effectiveness of Spatial Alignment Module.}
We validate the effectiveness of each components of SAM on the VR$\to$VID cross-domain experiment in Tab.~\ref{table:sam}. 
Firstly, we introduce the spatial alignment for positive and negative sample pairs, respectively. Each of them brings certain performance improvements. 
However, a unified alignment for all sample pairs damages ID-related features and degrades the performance instead. 
Therefore, we separate the alignment of positive and negative samples, which retains some discriminative features and integrates the effect of both aspects. 
The extra PAM for positive sample alignment further guarantees the focus on ID-related positions and achieves the best performance.
We also conduct in-domain experiment on VR to prove that SAM improves the generalization capability without sacrificing the supervision performance. 
Adding SAM to OSNet receives similar results, which validates the universality of SAM. 

\begin{table}[t]
\small
\centering
\caption{Supervised performance comparison between MSINet and Transformers on VR and MS datasets. }
\label{table:trans}
\begin{tabular}{lcccccc}
\toprule
    \multirow{2}{*}{Method} & \multirow{2}{*}{Params} & Inference & \multicolumn{2}{c}{MS} & \multicolumn{2}{c}{VR} \\
     & & Time & R-1$\uparrow$ & mAP$\uparrow$ & R-1$\uparrow$ & mAP$\uparrow$ \\
     \midrule
    DeiT-S\cite{he2021transreid} & $\sim$22M & 0.97x & 76.3 & 55.2 & 95.5 & 76.3 \\
    DeiT-B\cite{he2021transreid} & $\sim$86M & 1.79x & \textbf{81.9} & \textbf{61.4} & 95.9 & 78.4 \\
    ViT-B\cite{he2021transreid} & $\sim$86M & 1.79x & 81.8 & 61.0 & 96.5 & 78.2 \\
    \textbf{MSINet} & 2.3M & 0.71x & 81.0 & 59.6 & \textbf{96.8} & \textbf{78.8} \\
    \bottomrule
\end{tabular}
\end{table}

\begin{figure}[t]
    \centering
    \includegraphics[width=\columnwidth]{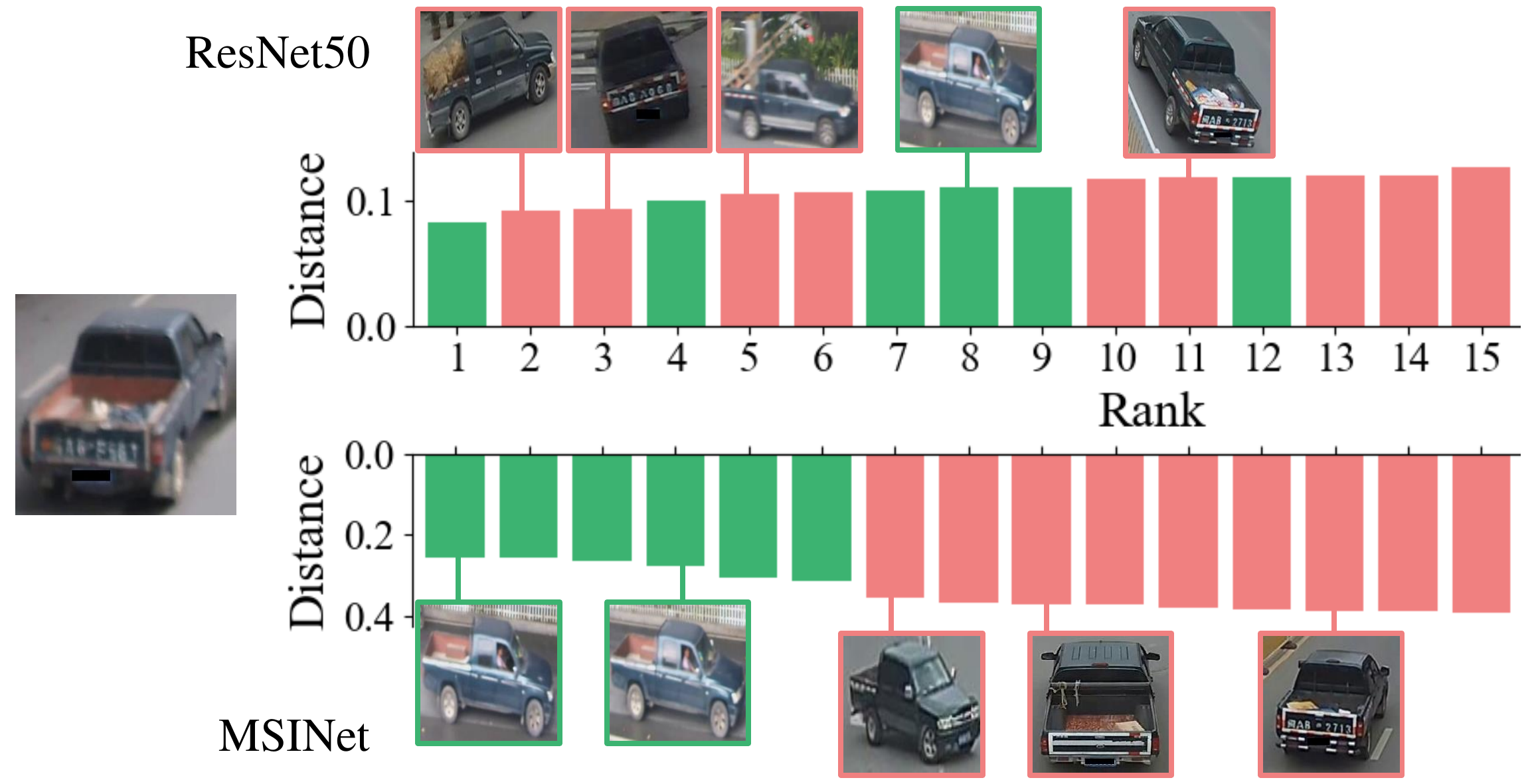}
    \caption{Example top-15 retrieved sequences comparison on VR. Appearance differences caused by variant camera conditions are well addressed by the proposed MSINet. Visualization of person ReID can be found in the supplementary material.}
    \label{fig:retrieval}
\end{figure}

{\bf Fusing Operation.}
After interaction, the multi-scale features are fused by sum operation. We investigate several fusing options on MS training from scratch in Fig.~\ref{fig:ablation} (c). Subtracting (``Minus'') a branch from the other leads to about the same results as ``Sum'' while multiplication (``Mul'') performs poorly. 

{\bf Comparison with Transformer.}
Transformer, as a new architecture, has recently been continuously making progresses in many computer vision domains~\cite{dosovitskiy2020image,touvron2021training}, including ReID~\cite{li2021diverse,he2021transreid}. We compare the model performance with some baseline Transformer models in Tab.~\ref{table:trans}. 
DeiT-B and ViT-B~\cite{he2021transreid} achieves higher performance on MSMT17, with much larger calculation burden compared with our proposed MSINet. On VeRi-776, MSINet surpasses all the baseline Transformer methods. 
It proves that rational interaction operations between multi-scale features are capable to assist light-weighted pure-CNN models to obtain comparable performance with complex Transformers. 

{\bf Parameter Analysis.}
Firstly, we study the influence of different receptive field scale ratios $\rho$ inside an MSI cell on MS training from scratch in Fig.~\ref{fig:ablation} (d).
Introducing scale differences between branches improves the model performance significantly, and subsequent increases brings more modest impacts. 
Considering both parameter scales and model performance, the ratio of 3:1 is selected for MSINet. 

Secondly, the model performance fluctuation influenced by spatial alignment weight is visualized in Fig.~\ref{fig:ablation} (e). The experiment is conducted on the VR$\to$VID cross-domain scenario. 
Employing the alignment generally makes a positive impact on the generalization capability of the model. The optimal loss weight $\lambda_{sa}$ locates at 2.0. 

{\bf Visualization Results.}
We visualize the top-15 retrieved sequences and the corresponding distances from an query image on VR in Fig.~\ref{fig:retrieval}. By comparison, ResNet50 mainly focuses on general appearance features, where the top-rated negative samples share similar car bodies. MSINet, oppositely, concentrates on discriminative distinctions, an empty car hopper in this case, and creates an evident distance gap between positive and negative samples. More details can be seen in the supplementation material. 

\section{Conclusion}
In this paper, we design a Twins Contrastive Mechanism for NAS to build the compatibility with ReID. The task-specific search scheme provides the searching process with more appropriate supervision. A Multi-Scale Interaction search space is proposed to establish rational and flexible utilization of multi-scale features. With a Spatial Alignment Module, our proposed MSINet achieves SOTA performance on both supervision and cross-domain scenarios with limited parameter amount. 
We hope the proposed approach could inspire more works focusing on designing network architectures suitable for ReID tasks. 

% \textbf{Limitations and Future Work}
% The designed interaction operations only include forward and exchange in the direct and attention forms, which restricts the size of the search space. 
% In the future works, there are still exploration room for more elaborate and complicated interaction operations and search spaces. 
% There are still exploration room for more complicated search schemes and spaces. 

\section{Acknowledgement}
This work is partially supported by National Natural Science Foundation of China (62173302, 62006244), China Scholarship Council (202206320302), Young Elite Scientist Sponsorship Program of China Association for Science and Technology (YESS20200140), and AI Singapore Programme (AISG2-PhD-2021-08-008).

%%%%%%%%% REFERENCES
{\small
\bibliographystyle{ieee_fullname}
\bibliography{egbib}
}

\clearpage

\section{Search on VeRi-776}

\setcounter{table}{6}
\begin{table}[t]
\small
\centering
\caption{The detailed interaction operation comparison between MSINet, MSINet-VR and MSINet-S. N: None; E: Exchange; G: Channel Gate; A: Cross Attention. }
\label{table:structure_comp}
\setlength{\tabcolsep}{4pt}
\begin{tabular}{l|c|c|c|c|c|c|c|c|c|c|c|c}
\toprule
    Model &\multicolumn{2}{c|}{\#1} & \multicolumn{2}{c|}{\#2} & \multicolumn{2}{c|}{\#3} & \multicolumn{2}{c|}{\#4} & \multicolumn{2}{c|}{\#5} & \multicolumn{2}{c}{\#6} \\
    \midrule
    \multirow{2}{*}{MSINet} & 1 & 2 & 3 & 4 & 5 & 6 & 7 & 8 & 9 & 10 & 11 & 12 \\
    & G & G & E & G & A & G & G & N & G & A & E & A \\
    \multirow{2}{*}{MSINet-VR} & 1 & 2 & 3 & 4 & 5 & 6 & 7 & 8 & 9 & 10 & 11 & 12 \\
    & G & G & E & A & G & A & G & A & A & A & E & A \\
    \multirow{2}{*}{MSINet-S} & 1 & 2 & 3 & 4 & 5 & 6 & 7 & 8 & 9 & 10 & 11 & 12 \\
    & E & A & G & A & A & A & G & A & E & A & E & A \\
    \bottomrule
\end{tabular}
\end{table}

\begin{table*}[t]
\small
\centering
\caption{Supervised performance on object ReID datsets. The results in the top part are trained from scratch, and those in the bottom part are pre-trained on ImageNet in advance. }
\label{table:supervised_comp}
\setlength{\tabcolsep}{2.8pt}
\begin{tabular}{lcccccccccccccc}
\toprule
    \multirow{2}{*}{Method} & \multirow{2}{*}{Params} & Inference & \multicolumn{2}{c}{M} & \multicolumn{2}{c}{MS} & \multicolumn{2}{c}{VR} & \multicolumn{2}{c}{VID} & \multicolumn{2}{c}{MS$\to$M} & \multicolumn{2}{c}{VR$\to$VID} \\
     & & Time & R-1$\uparrow$ & mAP$\uparrow$ & R-1$\uparrow$ & mAP$\uparrow$ & R-1$\uparrow$ & mAP$\uparrow$ & R-1$\uparrow$ & R-5$\uparrow$ & R-1$\uparrow$ & mAP$\uparrow$ & R-1$\uparrow$ & R-5$\uparrow$ \\
     \midrule
    ResNet50$^*$~\cite{luo2019strong} $\sim$ & 24M & 1x & 85.7 & 68.3 & 48.0 & 25.7 & 92.8 & 69.9 & 70.6 & 76.6 & - & - \\
    MSINet & 2.3M & 0.71x & 94.6 & 87.0 & \textbf{76.0} & \textbf{52.5} & \textbf{95.9} & 75.0 & \textbf{76.5} & \textbf{89.8} & - & - & - & - \\
    MSINet-VR & 2.3M & 0.71x & \textbf{94.7} & \textbf{87.4} & 75.5 & 51.8 & \textbf{95.9} & \textbf{76.0} & 75.2 & 86.3 & - & - & - & - \\
    \midrule
    ResNet50$^*$~\cite{luo2019strong} & $\sim$24M & 1x & 94.5 & 85.9 & 75.5 & 50.4 & 94.5 & 73.6 & 76.5 & 89.9 & 58.8 & 31.8 & 42.8 & 61.9 \\
    MSINet & 2.3M & 0.71x & 95.3 & \textbf{89.6} & \textbf{81.0} & \textbf{59.6} & 96.8 & \textbf{78.8} & \textbf{77.9} & \textbf{91.7} & \textbf{74.9} & \textbf{46.2} & 48.0 & 65.6 \\
    MSINet-VR & 2.3M & 0.71x & \textbf{95.4} & 89.0 & 80.1 & 57.5 & \textbf{97.0} & 78.6 & 77.3 & 91.3 & 72.9 & 44.8 & \textbf{48.5} & \textbf{66.2} \\
    \bottomrule
\end{tabular}
\end{table*}

\begin{table*}[t]
\centering
\caption{The evaluation results of models searched with different training-validation identity ratios on the MS dataset. * indicates that the model is searched with the softmax loss. }
\label{table:overlap}
% \rowcolors{4}{lightgray!50}{white}
\begin{tabular}{cccc|cccc}
\toprule
    \multicolumn{4}{c|}{w/o overlap} & \multicolumn{4}{c}{w/ overlap} \\
    \multirow{2}{*}{train (\%)} & \multirow{2}{*}{valid (\%)} & \multicolumn{2}{c|}{MS} & \multirow{2}{*}{train (\%)} & \multirow{2}{*}{valid (\%)} & \multicolumn{2}{c}{MS} \\
    &  & R-1$\uparrow$ & mAP$\uparrow$ &  &  & R-1$\uparrow$ & mAP$\uparrow$ \\
    \midrule
    90 & 10 & 70.6 & 46.2 & 60 & 60 & 75.3 & 51.2\\
    75 & 25 & 71.4 & 46.9 & 40 & 80 & 75.1 & 50.5 \\
    67 & 33 & 75.5 & 51.5 & \textbf{60} & \textbf{80} & \textbf{76.0} & \textbf{52.5} \\
    50 & 50 & 74.7 & 50.4 & 80 & 60 & 74.8 & 51.0 \\
    33 & 67 & 74.7 & 50.8 & 80 & 80 & 74.4 & 50.3 \\
    25 & 75 & 74.9 & 50.6 & 100 & 100 & 74.4 & 50.1 \\
    10 & 90 & 75.4 & 50.7 & 100* & 100* & 74.0 & 49.8 \\
    \bottomrule
\end{tabular}
\end{table*}

We select a training-validation ratio of 60\%-80\% in the searching process on MSMT17~\cite{wei2018person}. Without changing any specific configurations, we directly search for the rational interaction operations on VeRi-776~\cite{liu2016large,liu2016deep} dataset. The searched architecture is denoted as MSINet-VR. We compare the structure of MSINet and MSINet-VR in Tab~\ref{table:structure_comp}. Generally, the two searched architecture have common characteristics: Channel Gate is preferred in shallow layers, while Cross Attention is employed for more thorough information interaction in deep layers. 

Quantitatively, we also conduct relevant supervision and cross-domain experiments with MSINet-VR in Tab.~\ref{table:supervised_comp}. All the experiment configurations are kept the same as those of MSINet training. Although there are some fluctuations, generally MSINet-VR has similar performance to MSINet, and the retrieval accuracy still surpasses ResNet50~\cite{he2016deep,luo2019strong} by a large margin. 

\section{Search with Different Overlap Ratios}

With the identities of training and validation sets unbound, we conduct a series of experiments utilizing different data separation ratios in Tab.~\ref{table:overlap} to find the appropriate interaction operations for the network. 
Firstly, we separate the training and validation sets completely with no identity overlaps. It can be observed that a balanced train-validation ratio generally brings better performance. 
For the two extremes of data distribution, an over-small validation set makes the architecture optimizer stuck in local minima and achieves poor performance. On the contrary, an over-large validation set brings no severe damage to the architecture search process, despite that the model is still not optimal. It demonstrates that abundant validation data is essential for ReID NAS. 

\begin{figure}[t]
\centering
\includegraphics[width=\columnwidth]{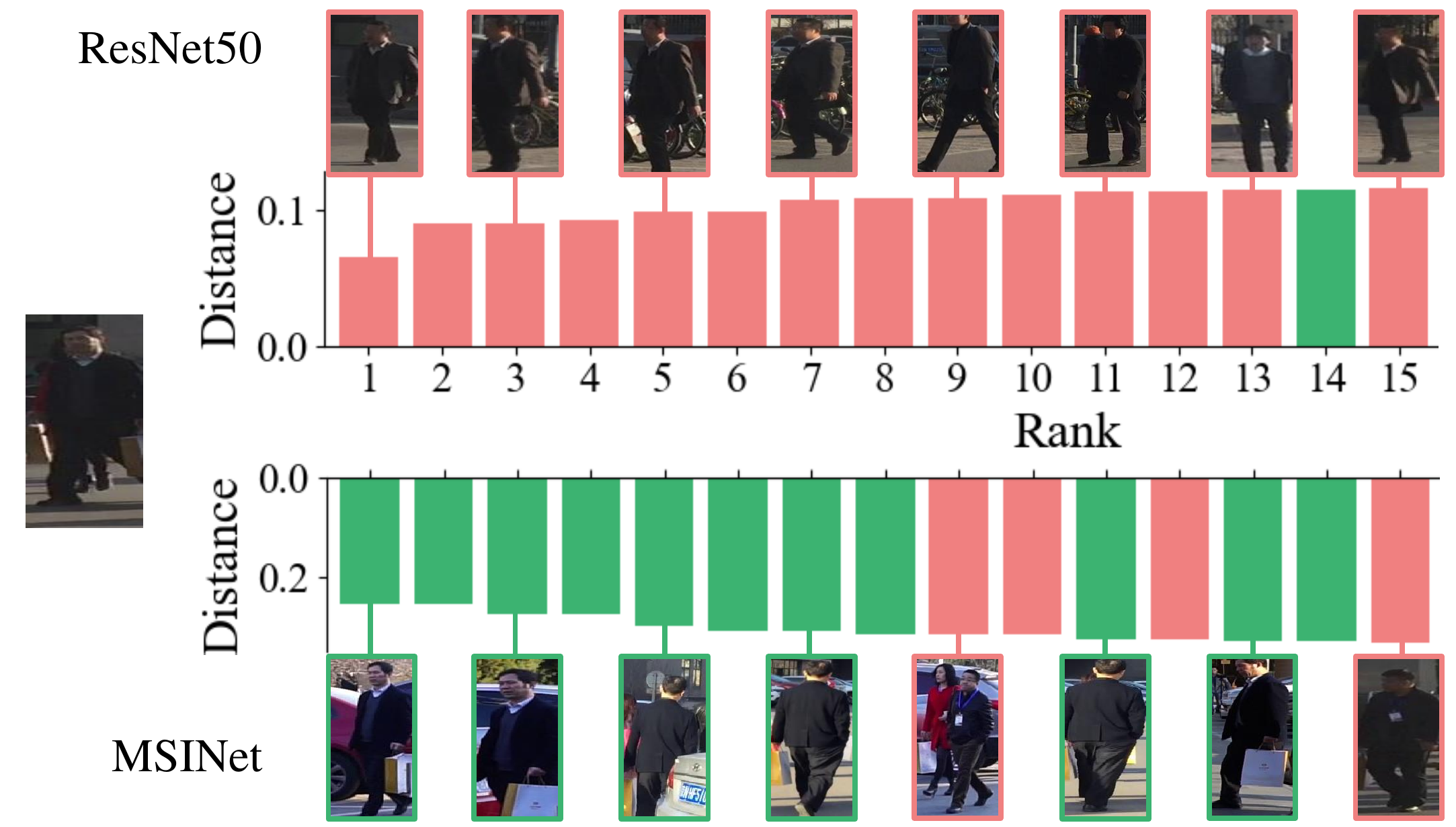}
\caption{Example top-20 retrieved sequences comparison on MSMT17. Best viewed in color. }
\label{fig:retrieval_vehicle}
\end{figure}

\begin{figure}[t]
\centering
\includegraphics[width=\columnwidth]{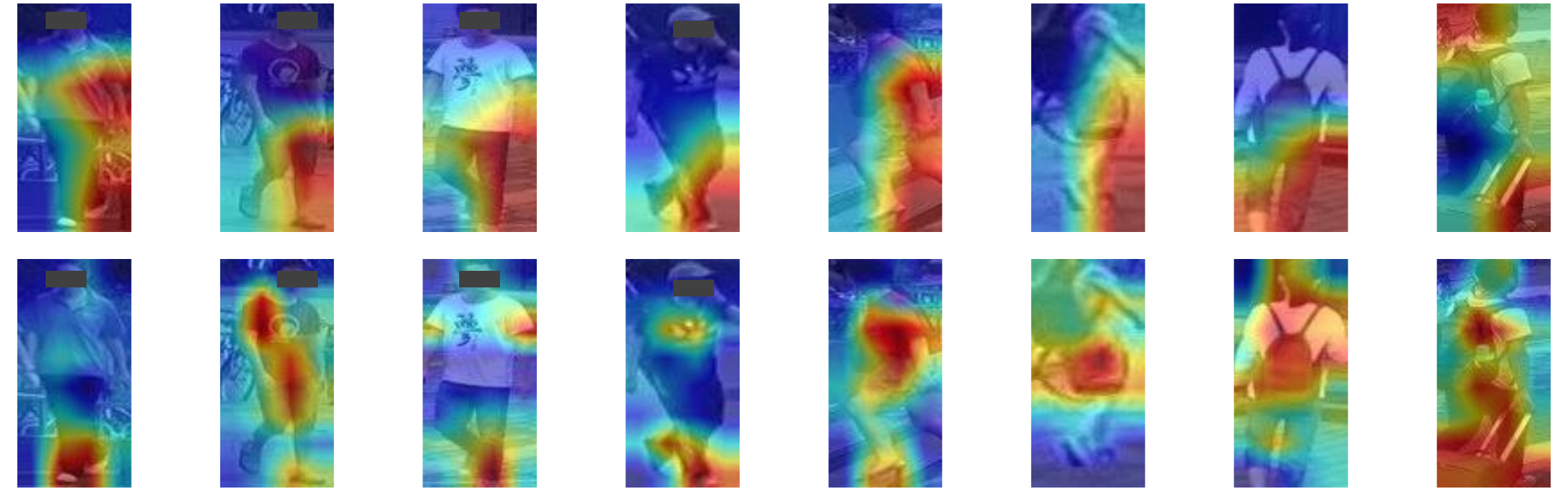}
\caption{Example activation maps of ResNet50 (the first row) and our proposed MSINet (the second row) trained on Market-1501 dataset. Best viewed in color. }
\label{fig:attention}
\end{figure}

Secondly, we randomly select part of identities, and evenly divide their images into the training and validation sets. 
The experiment results suggest that having a relatively small proportion of overlapped identities, whose images have been partly utilized for model parameter update, stabilizes the searching process and leads to a better architecture. However, when the overlap increases to a certain extent, the resemblance between the training and validation sets will bring negative influence to the ReID architecture search. 
As a comparison, we conduct the search task with traditional NAS scheme where a linear classification layer and cross entropy loss are employed for the training and validation data, the searched model of which performs worse than our proposed TCM.

Combined with above rules and the model performance, we select the architecture searched with the train-validation split of 60\%-80\% as the proposed MSINet.

\begin{figure}[t]
\centering
\includegraphics[width=\columnwidth]{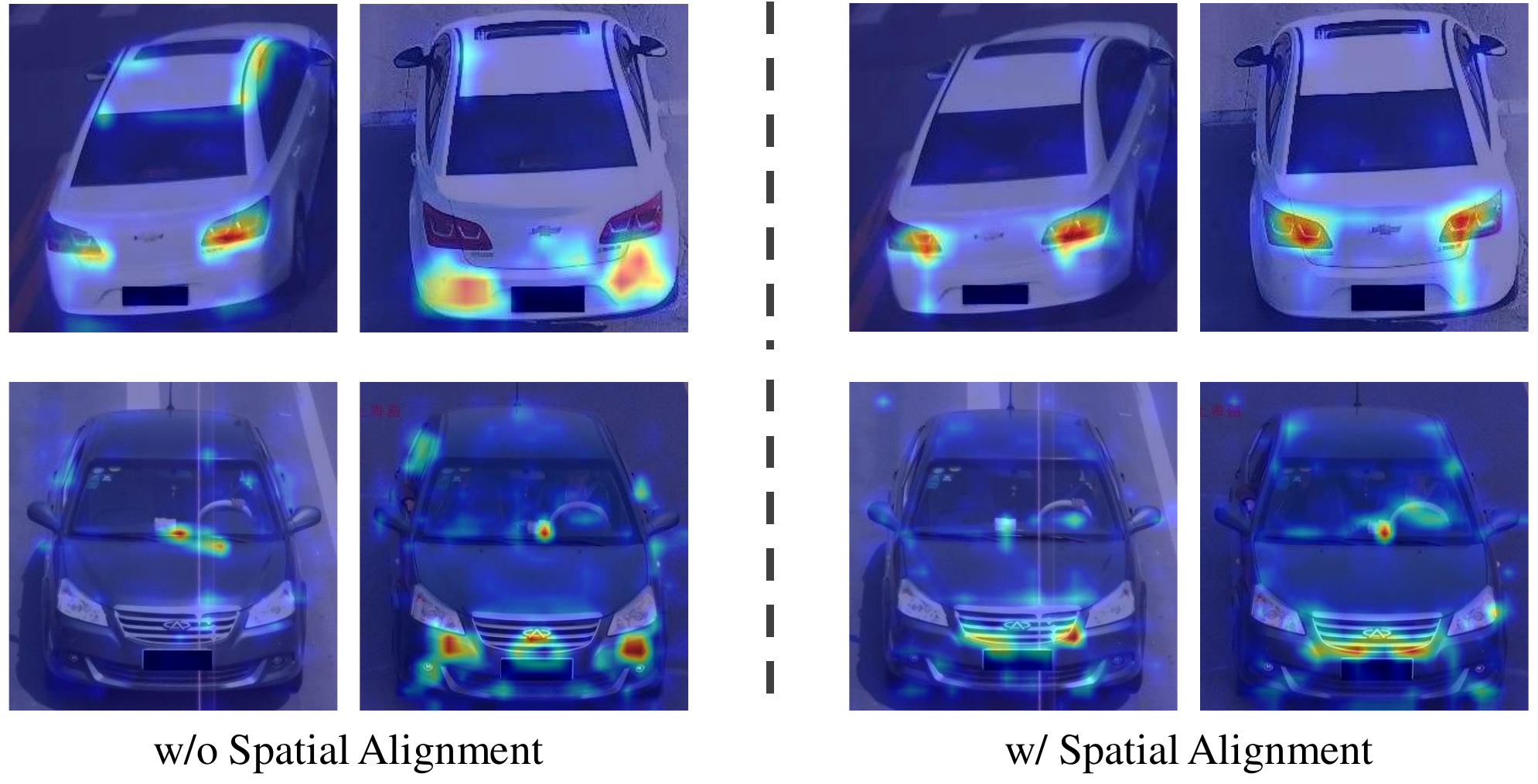}
\caption{Example activation maps of MSINet trained on the task of VR$\to$VID. With the Spatial Alignment Module, the model is capable to focus consistently on specific areas confronted with images from different sources. Best viewed in color. }
\label{fig:sam-attention}
\end{figure}

\section{Search with Softmax Loss}

We further compare the detailed interaction operations between MSINet and the architecture searched under traditional NAS scheme, where softmax loss and a unified linear classification layer are utilized for the training and validation sets~\cite{liu2018darts} (denoted as MSINet-S) in Tab.~\ref{table:structure_comp}. Compared with MSINet and MSINet-VR, where direct information exchange mainly appears at deep layers, MSINet-S contains a large amount of Exchange and Cross Attention along the whole network. The over-frequent information exchange fails to focus on discriminative features. It also validates the effectiveness of our proposed Twins Contrastive Mechanism on searching for architectures suitable for ReID. 

\section{Visualization Results}

Some additional visualization results are illustrated to further manifest the effectiveness of our proposed architecture. 
Firstly, we visualize an example comparison of the top-20 retrieved sequences between ResNet50 and MSINet on MSMT17 in Fig.~\ref{fig:retrieval_vehicle}. ResNet50 mainly focus on general appearance information, while our proposed MSINet concentrates on discriminative distinctions, the hand bag in this case. Even though positive samples have large appearance differences from the query image, MSINet is still capable to distinguish them. 

Secondly, example activation maps of ResNet50 and our proposed MSINet on Market-1501~\cite{zheng2015scalable} are visualized in Fig.~\ref{fig:attention}. ResNet50 mainly focuses on the right part of the image, including some background areas. Our proposed MSINet, oppositely, is capable to dynamically focus on discriminative distinctions of each image. 

Thirdly, to intuitively demonstrate the effectiveness of the Spatial Alignment Module (SAM) on enhancing the attention consistency of the model confronted with images from different sources, we visualize example activation maps on the task of VR$\to$VID. As shown in Fig.~\ref{fig:attention}, without alignment, the model can have different activated positions on different images of the same identity, even if they share similar appearances. 

\section{Comparison and advantages to OSNet}
(1) OSNet simply sums up the features of each branch, without detailed exploration on the interaction between branches. In comparison, MSINet practically select rational interaction operations for different network layers. Consequently, MSINet surpasses OSNet not only in supervised, but also in domain generalization performance by a large margin. 
(2) OSNet contains 4 branches with different receptive field scales, where there exists certain parameter redundancy. We validated in the early exploring that removing the branches with receptive field scales of 3 and 5 has little influence to the model performance. 
MSINet reduces the number of branches, and increases the scale difference between two branches, which increases the parameter amount by a little bit but significantly reduces the inference time. 

\section{Detailed Analysis on SAM}
We compare the proposed SAM module to some previous attention-based methods and analyze it in detail. 
\cite{zhou2019discriminative} regularizes the attention generated at different network layers for the same image; \cite{zheng2019re} explicitly enforces the longitudinal activation distribution to be the same for two images, which may lead to misalignment if the objects are not properly detected. 
For negative samples, there can be many different hints for recognition, some of which might be inappropriate, such as the backgrounds. By aggregating the information from different negative samples, the network is driven to only focus on discriminative regions. 
The motivation of SAM is different from the above two works. 
For the in-domain setting, the camera condition diffrences are directly addressed by supervised learning. 
Thus, SAM brings limited improvements, yet doesn't defect the performance, compared to techniques like instance normalization. 

\setlength{\tabcolsep}{10.5pt}
\begin{table}[t]
    \centering
    \small
    \begin{tabular}{lcccc}
    \toprule
    \multirow{2}{*}{Method} & \multicolumn{2}{c}{M$\to$MS} & \multicolumn{2}{c}{VID$\to$VR} \\
     & R-1$\uparrow$ & mAP$\uparrow$ & R-1$\uparrow$ & mAP$\uparrow$ \\
     \midrule
     OSNet & 21.2 & 7.2 & 69.2 & 32.0 \\
     MSINet & \textbf{22.4} & \textbf{8.3} & \textbf{72.1} & \textbf{33.8} \\
     \bottomrule
    \end{tabular}
    \vspace{-6pt}
    \caption{Additional Cross-domain Experiments}
    \label{tab:add_exp}
\end{table}

\setlength{\tabcolsep}{9pt}
\begin{table}[t]
    \centering
    \small
    \begin{tabular}{lcccc}
    \toprule
    \multirow{2}{*}{Method} & \multicolumn{2}{c}{M} & \multicolumn{2}{c}{VR$\to$VID} \\
     & R-1$\uparrow$ & mAP$\uparrow$ & R-1$\uparrow$ & R-5$\uparrow$ \\
     \midrule
     SAM & \textbf{95.5} & \textbf{89.9} & \textbf{49.0} & \textbf{66.8} \\
     SAM-softmax & 95.0 & 89.1 & 48.2 & 65.9 \\
     \bottomrule
    \end{tabular}
    \vspace{-6pt}
    \caption{Ablation study on softmax operation. }
    \vspace{-12pt}
    \label{tab:softmax_exp}
\end{table}

\section{More Ablation Study}
\textbf{Additional cross-domain experiments. }
We add the M$\to$MS and VID$\to$VR experiment results to Tab.~\ref{tab:add_exp}. MSINet surpasses OSNet on all metrics. 

\textbf{Ablation study on softmax operation. }
SAM aligns the activation values in the feature map, where the discriminative positions are actually matched between different samples. 
As the ``Mutual Conv'' operation is non-parametric, it is not proper to apply the softmax-squeezed position attention values for direct alignment, which may result in scale inconsistency. 
% Although it is the discriminative position that matters, the softmax-squeezed position attention values are not proper for direct alignment with the non-parametric activation from ``Mutual Conv''. 
The experiment results in Tab.~\ref{tab:softmax_exp} also suggest slight influence on this detail. 

\section{Limitations and Future Work}
The designed interaction operations only include forward and exchange in the direct and attention forms, which restricts the size of the search space. 
In the future works, there are still exploration room for more elaborate and complicated interaction operations and search spaces. 
There are still exploration room for more complicated search schemes and spaces. 

\end{document}